\documentclass[sn-mathphys,Numbered]{sn-jnl}

\usepackage{graphicx}%
\usepackage{multirow}%
\usepackage{amsmath,amssymb,amsfonts}%
\usepackage{amsthm}%
\usepackage{mathrsfs}%
\usepackage[title]{appendix}%
\usepackage{xcolor}%
\usepackage{textcomp}%
\usepackage{manyfoot}%
\usepackage{booktabs}%
\usepackage{algorithm}%
\usepackage{algorithmicx}%
\usepackage{algpseudocode}%
\usepackage{listings}%
\usepackage{array}
\usepackage{tabularx}
\usepackage{caption}
\usepackage{geometry}
\theoremstyle{thmstyleone}%
\theoremstyle{thmstyletwo}%

\theoremstyle{thmstylethree}%

\begin{document}

\title[Article Title]{SaudiBERT: A Large Language Model Pretrained on Saudi Dialect Corpora}


\author*[1]{\fnm{Faisal} \sur{Qarah}}\email{fqarah@taibahu.edu.sa}



\affil*[1]{\orgdiv{Department of Computer Science, College of Computer Science and Engineering}, \orgname{Taibah University}, \orgaddress{\city{Medina}, \postcode{42353}, \country{Saudi Arabia}}}




\abstract{

In this paper, we introduce SaudiBERT, a monodialect Arabic language model pretrained exclusively on Saudi dialectal text. To demonstrate the model's effectiveness, we compared SaudiBERT with six different multidialect Arabic language models across 11 evaluation datasets, which are divided into two groups: sentiment analysis and text classification. SaudiBERT achieved average F1-scores of 86.15\% and 87.86\% in these groups respectively, significantly outperforming all other comparative models. Additionally, we present two novel Saudi dialectal corpora: the Saudi Tweets Mega Corpus (STMC), which contains over 141 million tweets in Saudi dialect, and the Saudi Forums Corpus (SFC), which includes 15.2 GB of text collected from five Saudi online forums. Both corpora are used in pretraining the proposed model, and they are the largest Saudi dialectal corpora ever reported in the literature. The results confirm the effectiveness of SaudiBERT in understanding and analyzing Arabic text expressed in Saudi dialect, achieving state-of-the-art results in most tasks and surpassing other language models included in the study. SaudiBERT model is publicly available on \url{https://huggingface.co/faisalq/SaudiBERT}. }

\keywords{Transformers, Natural Language Processing, Distributed Computing, Monodialect Arabic Language Language Model, Artificial Intelligence}



\maketitle

\section{Introduction}\label{sec1}


Saudi Arabia holds a significant position on the global stage, not only as one of the world's largest oil producers but also as a spiritual center of the Islamic world, home to Makkah and Medina, the two holiest cities in Islam. Moreover, as Saudi Arabia executes its ambitious Vision 2030 \cite{vision2030}, which targets economic diversification and technological advancement, its influence is expanding into new sectors such as tourism, entertainment, and renewable energy. This sets the country as a transformative force in both the Middle East and the global economy.

The significant importance of Saudi Arabia highlights the need for developing advanced computational tools capable of handling the complexities of the Saudi dialect and its local variants. Analyzing and processing the Saudi dialect within the field of natural language processing (NLP) presents unique challenges due to its rich regional variants and linguistic nuances that distinguish it from Modern Standard Arabic (MSA) \cite{prochazka2013saudi}. Filled with local expressions, idioms, and accents that vary from one region to another, studying the features of the Saudi dialect sheds light on the social, cultural, and regional diversity of the Kingdom. This makes it a vital area of study for both linguistic and cultural research.

Despite its importance, computationally processing the Saudi dialect faces several challenges. One major obstacle is the absence of standardized spelling conventions, which results in it being written in various forms \cite{al2018suar}, leading to textual inconsistencies. Additionally, the dialect's informal style often includes slang and regional expressions, which further complicates the development of accurate computational tools \cite{alruily2020issues}. Such diversity demands the development of advanced computational models capable of understanding and processing both the contextual and linguistic aspects of the dialect effectively.

The motivation behind focusing on the Saudi dialect arises from its growing importance in digital communication and social media platforms where it is widely used. As digital platforms becoming more used for social interactions, the ability to accurately analyze and process the Saudi dialect becomes crucial for various applications such as sentiment analysis, social meaning extraction, and the detection of fake news and spam. This highlights the need for developing a robust language model that can effectively process and analyze text expressed in the Saudi dialect.


To address these challenges, in this paper, we introduce a new monodialectal large language model, SaudiBERT, based on the BERT architecture and pretrained exclusively on Saudi dialectal text. The proposed model is designed to effectively handle a wide range of analytic tasks, such as text classification and sentiment analysis that are expressed in the Saudi dialect. To demonstrate the performance of the new model, we evaluated SaudiBERT against six Arabic multidialectal language models across 11 evaluation datasets divided into two groups: sentiment analysis and text classification. SaudiBERT has achieved average F1-scores of 86.15\% and 87.86\% in these groups respectively, significantly outperforming all other comparative models.

Furthermore, SaudiBERT model was pretrained on two novel Saudi dialectal corpora: the Saudi Tweets Mega Corpus (STMC), which contains over 141 million tweets, and the Saudi Forums Corpus (SFC), which comprises 15.2 GB of text. To the best of our knowledge, these corpora are the largest of their kind ever reported in the literature.

This study not only contributes to the academic field by filling the gap in resources and tools focused on the Saudi dialect, but also has practical implications for technological progress in the region. We believe that SaudiBERT language model and the new corpora have potential for the future of Saudi dialect analysis, serving as valuable tools for a wide range of applications, including education, business, and social media analytics. Additionally, they offer significant benefits to researchers in the fields of linguistics and NLP.

\noindent \\The main contributions of this paper can be summarized as follows:

\begin{itemize}
\item Introducing SaudiBERT, a new language model pretrained exclusively on Saudi dialect, designed to effectively process and analyze Saudi dialect text.
\item Reporting state-of-the-art results in the majority of evaluation tasks expressed in Saudi dialect, outperforming other multidialectal Arabic language models. 
\item Constructing two novel Saudi dialectal corpora, which are the largest reported in literature.

\end{itemize}

The paper is organized as follows: Section 2 presents the related work. Section 3 introduces background about Transformers. In Section 4, we discuss the proposed model and the compiled corpora. The experimental procedure is presented in Section 5. The experimental results are discussed in Section 6. Finally, the conclusion is presented in Section 7.


\section{Related Work}\label{sec2}

\subsection{Large-Scale Saudi Dialectal Corpora}

In the early stages of Arabic NLP, where most research efforts have been focused on Modern Standard Arabic (MSA), numerous researchers have proposed large-scale multidialectal Arabic corpora to fill the gap of the absence of such corpora and to further improve the field of dialectal Arabic (DA) NLP. A significant portion of many of these corpora contains texts in the Saudi dialect. Mubarak and Darwish introduced the Multi-Dialectal Corpus of Arabic \cite{mubarak2014using} which was collected based on the geographical location of each tweet. Initially, the compiled corpus contains 175M Arabic tweets, and after several filtering stages which include selecting tweets based on certain dialectal words, it was refined to 6.5M tweets. Of these, 61\% are from Saudi Arabia while the remaining 39\% are tweets from other Arabic countries. The compiled corpus was used for training a dialect identification model, and has achieved over 93\% accuracy in identifying Saudi and Egyptian dialect texts.

Similarly, Khalifa et al. introduced the Gumar corpus \cite{khalifa2016large}, another large-scale multidialectal Arabic corpus for Arabian Gulf countries. The corpus consists of 112 million words (9.33 million sentences) extracted from 1200 novels that are publicly available and written in Arabian Gulf dialects, with 60.52\% of the corpus text being written in Saudi dialect.

In addition to multidialectal Arabic corpora, many researchers have presented various monodialect Saudi corpora. Tarmom et al. \cite{tarmom2020compression} introduced two monodialectal Arabic corpora: the Saudi Dialect Corpus (SDC) and the Egyptian Dialect Corpus (EDC) assembled from Facebook and Twitter for the purpose of detecting code-switching in Arabic text. Both corpora contain more than 210K words, with a total text size of 2MB each. Similarly, Alruily has presented the Dialectal Saudi Twitter Corpus \cite{alruily2020issues}, which contains 207K tweets written in Saudi dialect and collected from 101 Saudi Twitter users in 2017. 

Al-Twairesh et al. proposed the Saudi corpus for NLP Applications and Resources (SUAR) \cite{al2018suar} which was considered a pilot study to explore possible directions to facilitate the morphological annotation of the Saudi corpus. The new corpus is composed of 104K words collected from forums, blogs, and various social media platforms (Twitter, Instagram, YouTube, and WhatsApp). The corpus was automatically annotated using the MADAMIRA tool \cite{pasha2014madamira} and manually validated.

Another interesting Saudi dialect corpus is the SaudiWeb Novels Dataset (Rewayatech) presented by Addawood and Alzeer \cite{addawood2020rewayatech}. The Rewayatech corpus is a compilation of unpublished novels in Saudi dialect written by anonymous Saudi writers. The corpus contains 1267 novels that were collected between 2003 and 2015 from Graaam online forum\footnote{https://forums.graaam.com}.

Elgibreen et al. introduced the King Saud University Saudi Corpus (KSUSC) \cite{elgibreen2021incremental} for a serious attempt to create a large-scale Saudi dialect corpus in various domains to be used in future NLP tasks targeting Saudi dialect text. The proposed corpus is a compilation of 10 pre-existing publicly available corpora, in addition to text collected from various websites and social media platforms (YouTube, Twitter, and Facebook). The total amount of the compiled corpus is 184M sentences (1.238B words). However, according to the statistics presented by the authors, 89\% (164 million sentences) of KSUSC corpus consists of MSA text acquired from previously published corpora. Therefore, the actual Saudi dialect text comprises only a small fraction of the KSUSC corpus.

Based on the studies discussed above, it can be seen that the domain of the Saudi dialect corpora requires further contributions. Additionally, the current corpora's limitations in size are not enough for pretraining large language models such as BERT, which comes at the lower end compared to other language models in terms of size and requirements. The necessity for larger corpora becomes even more critical for models targeting monodialectal text, whether for generation or analysis purposes, especially since people are increasingly using dialectal Arabic. There is a significant need for the collection of more Saudi dialect text to improve the performance of Arabic language models targeting tasks contain such dialect-specific text. 

In summary, regardless of the rich literature on Saudi dialect corpora, a significant gap remains in terms of size and diversity, and Saudi dialect corpora are still lacking and need further contributions. Thus, in this paper we are proposing two new Saudi dialectal corpora specifically designed for pretraining large language models to improve the field of Saudi dialectal NLP.

\subsection{Multidialect and Monodialect Arabic LLM}

The rise of Arabic language models such as AraBERT \cite{antoun2020arabert}, ArabicBERT \cite{safaya2020kuisail}, AraGPT2 \cite{antoun2021aragpt2}, and ARBERT \cite{abdul2020arbert} has significantly advanced the field of Arabic NLP, especially for tasks related to Modern Standard Arabic (MSA). However, in experiments on dialectal Arabic (DA) tasks, these models have failed to achieve similar performance. Given these findings, numerous researchers have been encouraged to develop multidialectal and monodialectal models designed for tackling DA tasks.

Chowdhury et al. proposed QARiB \cite{chowdhury2020qarib} a BERT-based language model that was pretrained on both DA and MSA text. The pre-training data was a combination of Arabic GigaWord \cite{parker2011arabic}, Abulkhair Arabic Corpus \cite{1.5bwords}, OPUS \cite{lison2016opensubtitles2016}, and 420 Million tweets. The proposed model has achieved state of the art results compared to AraBERT on DA tasks presented by the authors. Similarly, Abdul-Mageed et al. \cite{abdul2020arbert} introduced two new BERT-based models called ARBERT and MARBERT. ARBERT was pretrained on 61 GB of MSA text collected from Arabic Wikipedia, online free books, and OSCAR \cite{oscar2019asynchronous}, whereas MARBERT was pretrained on a different dataset composed of one billion tweets written in both MSA and DA. Both models utilize a vocabulary size of 100k wordpieces and were pretrained using the same configuration as the original BERT. They have achieved state-of-the-art results on the majority of tasks when compared with AraBERT and other multilingual models.

Inoue et al. introduced four new BERT-based models called CAMeLBERT \cite{inoue2021camelbert}. Each variant was pretrained on a different type of Arabic text: Modern Standard Arabic (MSA), dialectal Arabic (DA), classical Arabic (CA), and a mixed of all three. The mixed variant was pretrained on a compilation of all datasets used in the first three variants totaling 167 GB of text, which included 54 GB of DA text compiled from 28 different Arabic dialectal corpora. When benchmarking CAMeLBERT-mix against existing models such as AraBERT, MARBERT, and ARBERT, the new model significantly outperformed all other models mainly on DA and CA tasks.

Antoun et al. \cite{antoun2020arabert} introduced a successor to the original AraBERT, named AraBERTv0.2, which was pretrained on a significantly larger dataset of 77 GB compared to the 22 GB dataset used in the pretraining of the original model. The new dataset was compiled from various corpora and public sources, including OSCAR \cite{oscar2019asynchronous}, OSIAN \cite{osian2019zeroual}, The 1.5B Arabic Words Corpus \cite{1.5bwords}, Arabic Wikipedia, and online news articles. Additionally, the authors presented an enhanced variant of the latter model called "AraBERTv0.2-Twitter" that was further pretrained on 60M DA tweets. Similarly, following the success of ARBERT and MARBERT, Elmadany et al. introduced ARBERTv2 and MARBERTv2 \cite{elmadany2023orca} that were pretrained on larger corpora than their predecessors and evaluated using the ORCA benchmark, where they demonstrated superior performance over previously proposed language models.

Additionally, numerous researchers have proposed monodialectal Arabic language models, including SudaBERT \cite{elgezouli2021sudabert} for the Sudanese dialect, TunBERT \cite{haddad2023tunbert} for the Tunisian dialect, DziriBERT \cite{abdaoui2021dziribert} for the Algerian dialect, and DarijaBERT \cite{gaanoun2024darijabert}, MorRoBERTa, and MorrBERT \cite{moussaoui2023pre} for the Moroccan dialect. However, these models were pretrained on relatively small corpora with sizes ranging from 67M to 691MB. Moreover, compared to other prominent Arabic language models they exhibit modest performance improvements on specific benchmarks. Similarly, Al-Yami and Al-Zaidy \cite{alyami2022weakly} developed seven Arabic RoBERTa models pretrained on a modest-sized dataset of Arabic tweets in various dialects (SA, EG, DZ, JO, LB, KU, and OM). These models were primarily designed for Arabic dialect detection and were compared with the original AraBERT and other multilingual language models. Among all the proposed models, AraRoBERTa-SA which was pretrained on the largest dataset (3.6M tweets) exhibited the highest accuracy in the benchmark used by the authors for detecting the Saudi dialect.

Based on the review of the literature presented above it can be seen that only a limited number of monodialect Arabic language models have been introduced, highlighting a significant gap in language models targeting specific dialects. Additionally, these monodialect models were pretrained on relatively small corpora, which may limit their effectiveness. Furthermore, among these models, only AraRoBERTa-SA was developed to target Saudi dialect tasks. However, as shown in the results section, the model underperforms when compared with other prominent multidialect models, suggesting that its pretraining corpus size is not enough to capture the linguistic nuances and semantics of the Saudi dialect. This demonstrates the need for the development of a Saudi dialect-specific model that is pretrained on a substantially larger corpus. Such a model would be crucial in advancing the field of Arabic NLP by significantly improving performance on tasks involving the Saudi dialect, thus addressing a significant gap in the existing language models.

In this study, all multidialect language models have been included in the analysis as comparative models. These include: AraBERTv02-Twitter, QARiB, CAMeLBERT-DA, MARBERTv1, MARBERTv2, and AraRoBERTa-SA. 

\section{Transformers and Bidirectional Encoder Representations from Transformers}\label{sec3}

\begin{figure}[h]
    \centering
    \includegraphics[width=0.5\textwidth]{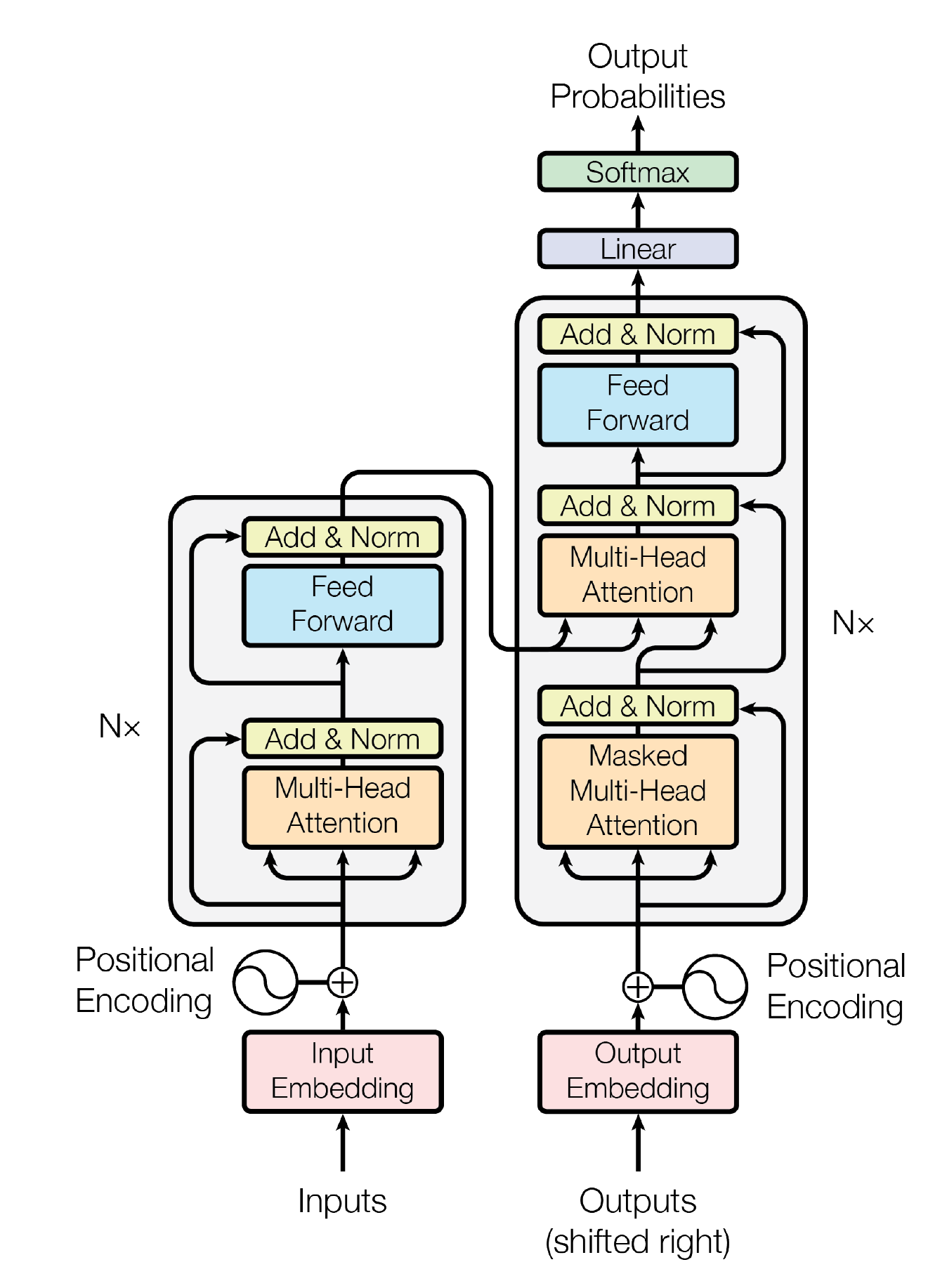}
    \caption{Overview of the transformer model architecture \cite{vaswani2017attention}}
    \label{fig1}
\end{figure}

Since the introduction of Transformers by Vaswani et al. \cite{vaswani2017attention} in 2017, all modern language models were built using this novel approach. They primarily utilize a mechanism called self-attention to measure the importance of different parts in the input text against each other. Figure \ref{fig1} shows the general architecture of Transformer model which contains a stack of two main components: encoders and decoders, where each utilize a paradigm called self-attention. 
The self-attention mechanism computes the weighted sum of the vectors for an input sequence using attention scores. The core components of the self-attention mechanism are the \textit{query} (Q), \textit{key} (K), and \textit{value} (V) matrices. These are acquired from the input vectors.

{}

The attention scores are computed as:
\begin{equation}
\text{Attention}(Q, K, V) = \text{softmax}\left(\frac{QK^T}{\sqrt{d_k}}\right) V
\end{equation}
where $d_k$ is the dimension of the key vectors.

While the self-attention mechanism allows the model to focus on different parts of the input, the multi-head attention mechanism allows the model to focus on different parts in different representation subspaces of Q, K, and V matrices simultaneously which improves the model understanding of the input sequence data. Essentially, it runs the self-attention mechanism multiple times in parallel, each with different learned linear projections of the original Q, K, and V.

Given $h$ different sets of Q, K, and V matrices, the multi-head attention is computed as:
\begin{equation}
\text{MultiHead}(Q, K, V) = \text{Concat}(\text{head}_1, \ldots, \text{head}_h) W_O
\end{equation}
where each head is computed as:
\begin{equation}
\text{head}_i = \text{Attention}(Q W_{Qi}, K W_{Ki}, V W_{Vi})
\end{equation}
and $W_{Qi}$, $W_{Ki}$, $W_{Vi}$, and $W_O$ are the parameter matrices.

However, since Transformers inherently lack a sense of order or position, the authors also proposed another mechanism called "positional encoding" that can give the model information about the position of words in a sequence, since all words or tokens are being processed in parallel. To address this, positional encodings are added to the embeddings at the bottoms of the encoder and decoder stacks. The positional encodings have the same dimension as the input embeddings, allowing them to be summed. The word's position \( p \) and each dimension \( i \) of the word embedding, the positional encoding is defined as:

The authors also introduced another mechanism known as "positional encoding", which is a technique used in Transformer models to provide them with information about the order or position of tokens in a sequence. Since the Transformer architecture processes all tokens in parallel and can not distinguish the order of these tokens by itself.
The positional encodings are calculated using the Equations \ref{sin_pos_encoding} and \ref{cos_pos_encoding}, and then added to the input embeddings before they are processed by the Transformer model. The positional encodings have the same dimension as the input embeddings, allowing them to be summed. The token's position \( p \) and each dimension \( i \) of the word embedding.
\begin{equation}
PE_{(p, 2i)} = \sin\left(\frac{p}{{10000^{2i/d}}}\right)
\label{sin_pos_encoding}
\end{equation}

\begin{equation}
PE_{(p, 2i+1)} = \cos\left(\frac{p}{{10000^{2i/d}}}\right)
\label{cos_pos_encoding}
\end{equation}

Where \( d \) is the dimension of the input embeddings. These sinusoidal functions were chosen because they can be easily learned if needed, and they allow the model to interpolate positions of tokens in long sequences.

Transformers original consist of encoders and decoders, where the encoder processes the input sequence and the decoder generates the output sequence. This architecture makes the original Transformer model particularly suitable for text-to-text tasks such as text-correction and machine translation tasks.

Devlin et al. \cite{devlin2018bert} proposed a new variant of the Transformer's architecture known as the Bidirectional Encoder Representations from Transformers (BERT). The new architecture has introduced significant advancements in field of natural language processing (NLP). BERT is an encoder-based Transformer that processes input text bidirectionally, unlike the original encoder-decoder Transformer model that can only process the input text sequentially. This has enabled BERT model to capture the complete context of a word or a token by considering its surrounding tokens. This is achieved through the "masked language model" (MLM) training objective, which randomly masks a set of tokens and then instructs the model to identify these masked tokens based on the context provided by the other unmasked tokens. 

\begin{figure}[]
    \centering
    \includegraphics[width=0.5\linewidth]{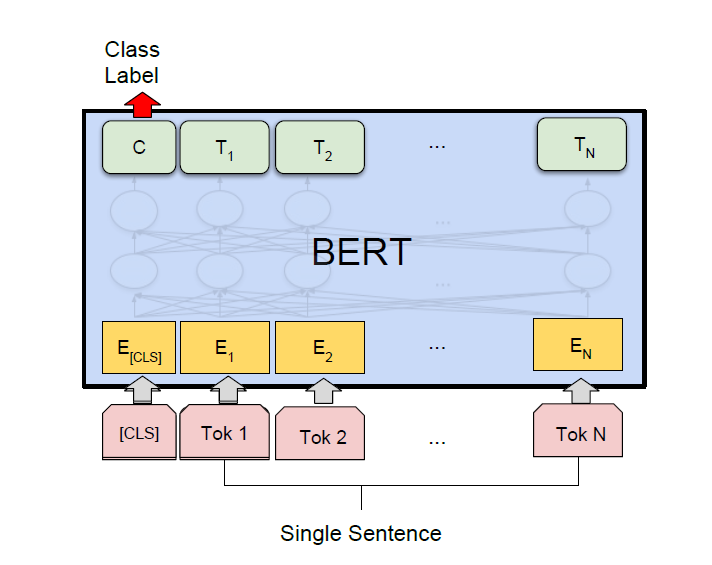}
    \caption{An illustration of BERT fine-tuning process for text classification \cite{devlin2018bert}.}
    \label{fig3}
\end{figure}

One of the main advantages of BERT is that it excels in transfer learning, which means utilizing previously learned knowledge on different NLP tasks such as sentiment analysis and question answering. After pretraining the model using a large corpus, it can be further fine-tunes by adding an extra output layer suitable for the target task. The fine-tuning process can be conducted on a basic hardware configuration without the need for significant modifications to the model architecture, which shows the high adaptability and efficiency of BERT language model. Figure \ref{fig3} shows the fine-tuning process of BERT for text classification tasks.


{}

\section{Methodology}\label{sec4}

In this section we describe the proposed model architecture, and the corpora used in pretraining the model.

\subsection{Saudi Corpora}
In this study we have compiled two new corpora that mainly contain Saudi dialectal texts: Saudi Tweets Mega Corpus (STMC) and Saudi Forums Corpus (SFC) to be used in pretrainig SaudiBERT model. To the best of our knowledge, both corpora are the largest Saudi dialectal corpora ever reported in literature.

\subsubsection{Saudi Tweets Mega Corpus}

The first corpus used in pretraining SaudiBERT was the Saudi Tweets Mega Corpus (STMC)\footnote{\url{https://huggingface.co/datasets/faisalq/STMC}}, compiled from a large in-house dataset of approximately 837 million Arabic tweets. These tweets were extracted based on the 'place' and 'location' metadata fields. The 'place' field contains information about the user's geolocation such as latitude, longitude, and the 'country-code'. We extracted tweets containing the value 'SA' in the country-code field indicating Saudi Arabia. However, only 2M tweets met this criterion, whereas the remaining tweets either associated with a different country-code or having an empty 'place' field.

For tweets lacking information in the 'place' field or belong to different country, we examined the text of the 'location' field. A significant portion of users mentioned their city or region, despite the majority providing information unrelated to their location. A comprehensive search was conducted for terms related to Saudi Arabia such as 'KSA', 'Saudi', the Saudi flag emoji, names of Saudi regions and cities, prominent Saudi soccer teams, and Saudi tribal names in both Arabic and English languages. In the search process we utilized regular expressions to examine if the content of the 'location' field contains any of the 187 Saudi-related terms that we compiled. However, the 'location' text required a considerable amount of cleaning and preprocessing to standardize the various writing styles used by the users. The text preprocessing includes eliminating consecutive letter repetitions, ignoring case sensitivity, removing diacritics, symbols, punctuation, and emojis (excluding the Saudi flag emoji), eliminating extensions (Tatweel) of Arabic letters, normalizing text by standardizing Arabic letter forms (e.g. informally 'Alif' can be written with or without 'Hamza'), and converting non-standard Latin letters to their English equivalents (e.g. á and à are converted to a). Through these measures, we retrieved more than 139 million tweets, resulting in a total corpus of 141,877,354 Saudi tweets. The STMC corpus is publicly accessible, but in compliance with Twitter's terms of service we have only released the tweet IDs. These IDs can be used to hydrate the tweets text using Twitter API.

Before adding it to the large text collection used in pretraining SaudiBERT, the STMC corpus went through few text preprocessing steps including the complete removal of URLs, user mentions, and hashtags, along with the elimination of newlines and any extra whitespaces. Additionally, all numbers larger than 7 digits were removed, and the repetition of letters was limited to five times, while other characters and emojis were allowed up to four repetitions. Tweets containing less than three words or those with more than 50\% of their text written in English are also removed. Finally, we eliminate all duplicate tweets to ensure the uniqueness of the text used in pretraining the model. 

Notably, all emojis, emoticons, punctuation, and diacritics were preserved, and the text was not subject to stop word removal, stemming, lemmatization, or any form of text normalization. This approach ensures the preservation of the original text's full semantics and enhances the model's ability to capture the nuances of informal Arabic text. After applying all preprocessing steps, the final text size of the STMC corpus is 11.1 GB, containing 1,174,421,059 words and 99,191,188 sentences.


\subsubsection{Saudi Forums Corpus}

The second corpus used for pretraining SaudiBERT was the Saudi Forums Corpus (SFC) which was compiled from five different Saudi online forums: Btalah\footnote{\url{https://www.btalah.com/}}, Hawamer\footnote{\url{https://hawamer.com/vb/index.php}}, Kooora\footnote{\url{https://www.kooora.com/}}, Mbt3th\footnote{\url{https://www.mbt3th.us/vb/forum.php}}, and Mekshat\footnote{\url{https://mekshat.com/vb/}}. The scraping process started by downloading all the HTML files from certain sections of each forum. After that we used BeautifulSoup API \cite{richardson2007beautiful} to extract the main post and the replies from these files locally.

Once all text was extracted, we applied the same preprocessing steps used on the STMC corpus to ensure the quality of the text before being used for pretraining the model. This included removing URLs, email addresses, newlines and extra whitespaces, and all numbers larger than 7 digits. We also limited the repetition of letters to five and other characters to four. Texts with less than three words or those with more than 50\% of their content written in English were also removed.

Additionally, we removed all BBCode tags and their attributes, HTML tags, percent-encoded sequences, and all duplicate texts. The final size of the SFC corpus is 15.2 GB, containing 1,532,752,665 words and 70,859,216 sentences. 
The complete SFC corpus will not be made public. However, we are releasing a subset called SFC-mini\footnote{\url{https://huggingface.co/datasets/faisalq/SFC-mini}}, which represents 20\% of the full corpus that was randomly sampled. SFC-mini contains 306,400,210 words and 14,181,729 sentences.





\subsection{SaudiBERT Architecture and Pretraining Procedures}

SaudiBERT is a BERT-based language model that was pretrained exclusively on Saudi dialectal text from scratch. The model follows the same architecture as the original BERT model with 12 encoder layers, 12 attention heads per layer, and a hidden layer size of 768 units. However, unlike the original BERT which used the WordPiece tokenizer \cite{wu2016google}, we employed the SentencePiece tokenizer \cite{kudo2018sentencepiece} for tokenizing the pretraining corpora, following the approach of more recent language models such as DeBERTaV2 \cite{he2021deberta}, ALBERT \cite{lan2019albert}, and XLM-R \cite{conneau2020unsupervised}. Additionally, we set the vocabulary size of SaudiBERT model to 75,000 wordpieces, enabling it to capture a wide range of terms and expressions found in Saudi dialectal text, including emojis. 

Traditionally, BERT language models are usually pretrained on two training objective tasks: the masked language model (MLM), and the next sentence prediction (NSP). In the MLM task, a proportion of the input tokens are masked and the model's objective is to accurately predict these masked tokens. In the NSP task, the model is provided with two sentences and must determine whether they are related or not. However, SaudiBERT was pretrained exclusively on the MLM training objective task, which reduces the model pretraining time and could potentially enhances the model performance in downstream tasks \cite{liu2019roberta}. 

SaudiBERT model was pretrained on the newly compiled SFC and STMC corpora, which together have a total text size of 26.3GB. The training was conducted on a single Nvidia 80GB-H100 SXM GPU provided by a cloud GPU rental service, using the Huggingface Transformers library \cite{wolf2020transformers} with the following configurations: masking 15\% of the input tokens, a maximum sequence length of 128, a batch size of 256, and the 'AdamW' optimizer with a learning rate of 5e-5. To optimize GPU memory usage and reduce training time, we used the mixed precision data type "FP16" for gradient computations. With these settings, the model was pretrained for 12 epochs (3.08M steps, 342 hours) and achieved a training loss of 2.38.

\section{Experiments}\label{sec5}

In this section, we describe the evaluation tasks and the fine-tuning  procedures. 
All experiments were conducted on a local machine with an AMD Ryzen-9 7950x processor, 64GB of DDR5 memory, and two GeForce RTX 4090 GPUs, each with 24GB of memory. We set up our software environment on Ubuntu 22.04 operating system and used CUDA 11.8 with Huggingface transformers library to download and fine-tune the comparative language models from the Huggingface hub along with our proposed model.

\subsection{Fine-tuning Procedures}
In this study, we compared the performance of SaudiBERT against six different Arabic dialectal language models: AraBERTv02-Twitter\footnote{\url{https://huggingface.co/aubmindlab/bert-base-arabertv02-twitter}}, QARiB\footnote{\url{https://huggingface.co/qarib/bert-base-qarib}}, CAMeLBERT-DA\footnote{\url{https://huggingface.co/CAMeL-Lab/bert-base-arabic-camelbert-da}}, MARBERTv1\footnote{\url{https://huggingface.co/UBC-NLP/MARBERT}}, MARBERTv2\footnote{\url{https://huggingface.co/UBC-NLP/MARBERTv2}}, and AraRoBERTa-SA\footnote{\url{https://huggingface.co/reemalyami/AraRoBERTa-SA}}. All models were fine-tuned using identical hyperparameters for all evaluation tasks: a maximum sequence length of 128, a batch size of 64, 'AdamW' optimizer with a learning rate of 5e-5, and the mixed precision data type "FP16" for gradient computations.
In all experiments, the models were evaluated using F1-score and Accuracy metrics. Additionally, each experiment was conducted three times to ensure reliable results, and we reported the highest F1-score achieved by each model on every evaluation task. It is important to note that the number of epochs and validation steps varies for each task depending on the number of samples and the complexity of each dataset.

For each task, the dataset was divided into 80\% for training and the remaining 20\% reserved for validation, while the same validation set was used for evaluating different models within the same experiment.

\subsection{Evaluation Tasks}
To exhibit the performance of SaudiBERT model, we evaluated its performance with six comparative models on two groups of downstream tasks. The sentiment analysis group contains six tasks, whereas the text classification group contains five tasks.

\subsubsection{Sentiment Analysis}

Sentiment analysis has been a popular research topic in the field of Arabic NLP, with numerous datasets and approaches proposed in the literature \cite{alqurashi2023arabic}\cite{almutairi2023comparative}. In this study, more than half of the evaluation tasks are based on datasets specifically designed to address sentiment analysis in the Saudi dialect. Table \ref{tab1} provides a detailed summary of the sentiment analysis evaluation tasks.

\begin{table}[h]
\centering
\caption{Summary of sentiment analysis datasets}
\label{tab1}
\begin{tabular}{lclcccp{3.5cm}}
\hline
\textbf{Dataset} & \textbf{Reference} & \textbf{Type} & \textbf{\#Labels} & \textbf{Train Size} & \textbf{Validation Size} & \textbf{Link} \\ \hline
AraCust & NA & Sentiment Analysis & 2 & 16k & 4k & \url{https://peerj.com/articles/cs-510/#supplemental-information} \\ 
SS2030 & NA & Sentiment Analysis & 2 & 3.4k & 0.85k & \url{https://www.kaggle.com/datasets/snalyami3/arabic-customer-reviews} \\ 
SSI-Electronics & NA & Sentiment Analysis & 3 & 1.6k & 0.4k & \url{https://ieee-dataport.org/documents/saudishopinsights-electronics} \\ 
SSI-Clothes & NA & Sentiment Analysis & 3 & 1.71k & 0.42k & \url{https://ieee-dataport.org/documents/saudishopinsights-clothes} \\ 
Saudi Bank Sentiment & NA & Sentiment Analysis & 3 & 9.63k & 2.4k & \url{https://github.com/iwan-rg/Saudi-Bank-Sentiment} \\ 
SDCT-Sentiment & NA & Sentiment Analysis & 3 & 3.34k & 0.83k & Directly from the author \\ \hline
\end{tabular}
\end{table}

\begin{itemize}
    \item 
    \textbf{AraCust:} \\
    AraCust \cite{almuqren2021aracust} is the first Arabic gold standard dataset for analyzing sentiments of Saudi telecommunication customers. It consists of tweets collected between January and June 2017 from eight Twitter accounts related to the main Saudi telecom companies (STC, Mobily, and Zain). Initially, the authors collected 3.5 million tweets, but after careful filtering and cleaning, the dataset size was reduced to 20,000 tweets written in Saudi dialect. These tweets were manually labeled as either "positive" or "negative," with negative sentiments comprising 68\% of the dataset and positive sentiments 32\%.
    
\vspace{2mm}

    \item 
    \textbf{SS2030:} \\
    The SS2030 dataset \cite{alyami2020application} was built from tweets written in Saudi dialect discussing various issues and events related to Saudi Vision 2030. The dataset is composed of 4,200 tweets that have been manually annotated as either 'positive' or 'negative'.
    
\vspace{2mm}

    \item 
    \textbf{SaudiShopInsights Dataset (SSI):} \\
    The authors of the SaudiShopInsights Dataset (SSI) \cite{6e56-4e15-23} have introduced two new datasets composed of customer reviews expressed in Saudi dialect that were collected from various online shopping platforms. These reviews cover items in two categories: electronics and clothing. Each dataset is annotated as 'positive', 'negative', or 'neutral'. The SSI-Electronics dataset contains 2,000 reviews, whereas the SSI-Cloths dataset includes 2,142 reviews. 
    
\vspace{2mm}

     \item 
    \textbf{Saudi Bank Sentiment:} \\
    Another benchmark based on Twitter is the Saudi Bank Sentiment dataset \cite{alqahtani2022customer}. The authors collected 12,300 tweets in Saudi dialect from users expressing their opinions on four Saudi banks: AlRajhi, Alinma, Saudi National Bank (SNB), and Saudi Investment Bank (SAIB). The tweets were collected from the official Twitter accounts of the target banks throughout the month of September 2021. The collected data were manually annotated with one of three sentiments: positive, negative, and neutral, with negative sentiments comprising 70\% of the dataset whereas positive and neutral sentiments 30\%.
    
\vspace{2mm}

     \item 
    \textbf{SDCT:} \\
    The Multi-Dialects Corpus Classification for Saudi Tweets (SDCT) \cite{bayazed2020sdct} dataset consists of 4,181 tweets collected from various trending hashtags in Saudi Arabia. It is manually annotated for two different NLP tasks: sentiment analysis and Saudi local-dialect identification. For sentiment analysis, the tweets are categorized into one of three sentiments: positive, negative, or neutral. Additionally, each tweet is labeled according to one of the three main Saudi local dialects—Hijazi, Najdi, and Eastern—using distinguishing keywords for each dialect. In this study, the SDCT dataset is utilized for both evaluation task groups. SDCT-Sentiment is included in the sentiment analysis group, and SDCT-Dialect-Identification is part of the text classification group.

\end{itemize}

\subsubsection{Text Classification}

The text classification group contains a variety of NLP tasks expressed in Saudi dialect. Available tasks in this group include event detection, author's gender identification, sarcasm detection, Saudi dialect identification, and identification of specific Saudi local dialects. The last task is described in the SDCT dataset, while the other tasks are described below. Table \ref{tab2} presents a brief summary of the text classification evaluation tasks.

\begin{table}[h]
\centering
\caption{Summary of various datasets included in the text classification group}
\label{tab2}
\begin{tabular}{lcp{3cm}cccp{3.5cm}}
\hline
\textbf{Dataset} & \textbf{Reference} & \textbf{Type} & \textbf{\#Labels} & \textbf{Train Size} & \textbf{Validation Size} & \textbf{Link} \\ \hline
SDCT-Dialect-Ident & NA & Dialect Identification & 3 & 3.34k & 0.83k & Directly from the author \\ 
SDC-EDC & NA & Dialect Identification & 2 & 22.89k & 5.72k & \url{https://github.com/TaghreedT} \\ 

SDTwittC & NA & Gender Identification & 2 & 360k & 90k & \url{https://ieee-dataport.org/documents/saudi-dialect-corpus} \\ 
FloDusTA & N & Event Detection & 4 & 7.2k & 1.8k & \url{https://github.com/BatoolHamawi/FloDusTA} \\ 
Saudi Irony & NA & Sarcasm Detection& 2 & 15.7k & 3.92k & \url{https://github.com/iwan-rg/Saudi-Dialect-Irony-Detection} \\ \hline
\end{tabular}
\end{table}

\begin{itemize}
    \item 
    \textbf{Saudi Irony:} \\
    The Saudi Irony dataset (Sa'7r) \cite{almazrua2022sa} was designed to enhance various machine learning and language models for the task of sarcasm detection. It includes 19,810 tweets that are manually annotated as either 'ironic' or 'non-ironic'.

\vspace{2mm}

    \item 
    \textbf{FloDusTA:} \\
    The Flood, Dust Storm, Traffic Accident (FloDusTA) dataset \cite{hamoui2020flodusta} serves as a benchmark for emergency event detection. It comprises a collection of 9,000 tweets collected from March to September 2018, which are manually categorized into one of four classes: Flood, Dust Storm, Traffic Accident, or non-event. 52\% of the dataset are classified as 'non-event,' while the remaining 48\% are distributed among the emergency event classes.

\vspace{2mm}

    \item 
    \textbf{SDTwittC:} \\
    The Saudi Dialect Twitter Corpus (SDTwittC) \cite{alanazi2019toward} is a collection of a large number of tweets meant for identifying the tweet writer's gender. The author collected tweets from 200 Saudi Twitter users (100 users for each gender), with 233,926 tweets written by males, and 219,740 by females. In this study, SDTwittC is considered the largest dataset used in the evaluation process with a collection of 450,000 tweets and text size of 36.8 MB.
    
\vspace{2mm}

    \item 
    \textbf{SDC-EDC:} \\
    Tarmom et al. \cite{tarmom2020compression} introduced two new Arabic dialectal corpora: the Saudi Dialect Corpus (SDC) and the Egyptian Dialect Corpus (EDC). The SDC contains 14,891 comments and tweets collected from Facebook and Twitter platforms, whereas the EDC includes 13,739 comments mainly collected from Facebook. We created a new benchmark from these two corpora by labeling each with its designated dialect and then combining them. The purpose of this task is to evaluate different language models on their ability to distinguish between Saudi and Egyptian dialectal texts.


\end{itemize}

\section{Results}\label{sec6}

In this section we present the evaluation results\footnote{the code files for models finetuning along with the evaluation results are available on \url{https://github.com/FaisalQarah/SaudiBERT}} of all language models on the validation set of each task using the evaluation metrics 'Accuracy' and 'Macro F1-score'. For the sentiment analysis evaluation group, Table \ref{tab3} shows that SaudiBERT model has achieved the highest F1-score across all tasks, and the highest Accuracy score in four out of six tasks. Overall, SaudiBERT has significantly outperformed all other models in the sentiment analysis group, with an average Accuracy of 90.06\% and an average F1-score of 86.15\%. It is followed by MARBERTv2 (M6) which has average scores of 89.53\% for Accuracy and 85.18\% for F1-score respectively.

\begin{table}[h]
\centering
\caption{Evaluation results of various models on sentiment analysis datasets. Model labels: M2: AraBERTv0.2-Twitter, M3: QARiB, M4: CAMeLBERT-DA, M5: MARBERTv1 M6: MARBERTv2 M7: AraRoBERTa-SA. All results are rounded to two decimal places. The average score is calculated as the unweighted average of all scores.}
\label{tab3}
\begin{tabular}{lcccccccccccccccc}
\hline
 & \multicolumn{2}{c}{\textbf{SaudiBERT}} & \multicolumn{2}{c}{\textbf{M2}} & \multicolumn{2}{c}{\textbf{M3}} & \multicolumn{2}{c}{\textbf{M4}} & \multicolumn{2}{c}{\textbf{M5}} & \multicolumn{2}{c}{\textbf{M6}} & \multicolumn{2}{c}{\textbf{M7}} \\ 
\hline
 \textbf{Dataset}& \textbf{Acc.} & \textbf{F1} & \textbf{Acc.} & \textbf{F1} & \textbf{Acc.} & \textbf{F1} & \textbf{Acc.} & \textbf{F1} & \textbf{Acc.} & \textbf{F1} & \textbf{Acc.} & \textbf{F1} & \textbf{Acc.} & \textbf{F1} \\ 
\hline
AraCust &  97.87 &  \textbf{97.55} & 97.87 & 97.51 & 97.82 & 97.45 & \textbf{97.90} & 97.53 & 97.87 & \textbf{97.55} & 97.75 & 97.41 & 97.62 & 97.22 \\
SS2030 &  \textbf{95.53} &  \textbf{95.43} & 95.06 & 94.97 & 94.59 & 94.51 & 92.00 & 91.85 & 93.89 & 93.75 & 94.71 & 94.56 & 88.13 & 88.01 \\
SDCT-Sentiment &  \textbf{76.22} &  \textbf{73.09} & 72.64 & 68.28 & 73.11 & 68.88 & 71.32 & 65.46 & 75.98 & 71.28 & 73.35 & 70.42 & 63.32 & 57.85 \\
SSI-Electronics &  \textbf{95.55} &  \textbf{95.42} & 94.75 & 94.71 & 94.75 & 94.71 & 92.75 & 92.62 & 94.50 & 94.41 & 95.25 & 95.20 & 90.00 & 89.84 \\
SSI-Clothes & 88.31 &  \textbf{82.18} &  \textbf{90.88} & 77.38 & 89.25 & 80.50 & 89.01 & 75.14 & 88.08 & 76.38 & 89.95 & 80.58 & 84.81 & 68.06 \\
Saudi Bank  &  \textbf{86.93} &  \textbf{73.23} & 84.06 & 71.36 & 83.65 & 70.49 & 83.32 & 70.55 & 85.06 & 71.21 & 86.18 & 72.93 & 83.77 & 69.18 \\
\hline
 Avg. & \textbf{90.06} & \textbf{86.15}& 89.21 & 84.04 & 88.86 & 84.42 & 87.72 & 82.19 & 89.23 & 84.10 & 89.53 & 85.18 & 84.61 & 78.36\\ \hline
\end{tabular}
\end{table}

\begin{table}[h]
\centering
\caption{Evaluation results of various models on the datasets included in the text classification group. Model labels: M2: AraBERTv0.2-Twitter, M3: QARiB, M4: CAMeLBERT-DA, M5: MARBERTv1 M6: MARBERTv2 M7: AraRoBERTa-SA. All results are rounded to two decimal places. The average score is calculated as the unweighted average of all scores.}
\label{tab4}
\begin{tabular}{lcccccccccccccccc}
\hline
 & \multicolumn{2}{c}{\textbf{SaudiBERT}} & \multicolumn{2}{c}{\textbf{M2}} & \multicolumn{2}{c}{\textbf{M3}} & \multicolumn{2}{c}{\textbf{M4}} & \multicolumn{2}{c}{\textbf{M5}} & \multicolumn{2}{c}{\textbf{M6}} & \multicolumn{2}{c}{\textbf{M7}} \\ 
\hline
\textbf{Dataset} & \textbf{Acc.} & \textbf{F1} & \textbf{Acc.} & \textbf{F1} & \textbf{Acc.} & \textbf{F1} & \textbf{Acc.} & \textbf{F1} & \textbf{Acc.} & \textbf{F1} & \textbf{Acc.} & \textbf{F1} & \textbf{Acc.} & \textbf{F1} \\ 
\hline
Saudi Irony & \textbf{71.78} & \textbf{70.45} & 70.69 & 69.67 & 71.40 & 70.29 & 70.10 & 69.58 & 70.91 & 69.28 & 70.97 & 69.76 & 70.25 & 68.82 \\
SDC-EDC & 98.00 & 98.00 & 97.58 & 97.58 & 97.79 & 97.79 & 97.27 & 97.26 & 97.62 & 97.62 & \textbf{98.02} & \textbf{98.02} & 96.38 & 96.37 \\
SDCT-Dialect-Ident. & \textbf{98.08} & \textbf{98.08} & 94.98 & 95.02 & 97.73 & 97.75 & 96.29 & 96.36 & 97.73 & 97.72 & 97.73 & 97.72 & 94.86 & 94.95 \\
SDTwittC & \textbf{76.66} & \textbf{76.58} & 73.87 & 73.84 & 73.32 & 73.32 & 74.28 & 74.22 & 74.40 & 74.15 & 75.82 & 75.71 & 72.58 & 72.46 \\
FloDusTA & \textbf{96.33} & \textbf{96.20} & 96.22 & 96.02 & 95.61 & 95.37 & 94.61 & 94.43 & 95.66 & 95.58 & 96.16 & 96.00 & 93.33 & 92.99 \\ \hline
Avg. &  \textbf{88.17} &  \textbf{87.86} & 86.67 & 86.42 & 87.17 & 86.90 & 86.51 & 86.37 & 87.26 & 86.87 & 87.74 & 87.44 & 85.48 & 85.12 \\ \hline

\end{tabular}
\end{table}

For the text classification task group, Table \ref{tab4} shows that SaudiBERT has achieved the highest Accuracy and F1-score in four out of five tasks, with being slightly behind MARBERTv2 (M6) in the SDC-EDC task which contains a large number of text written in the Egyptian dialect. Overall, SaudiBERT has significantly outperformed all comparetive models in the text classification group, with an average Accuracy of 88.17\% and an average F1-score of 87.86\%. MARBERTv2 (M6) was the second best model with an average Accuracy of 87.74\% and an average F1-score of 87.44\%

Furthermore, the size of SaudiBERT model is 143M parameters, which is 12\% smaller than MARBERTv2 language model with size of 163M parameters. This reduction is due to the smaller vocabulary size of SaudiBERT, which contains 75k wordpieces compared to MARBERTv2's 100k wordpieces. Additionally, SaudiBERT required significantly less pretraining time (12 epochs) compared to MARBERTv2, which was pretrained for 40 epochs on a 160 GB corpus.

\section{Conclusion}\label{sec7}

In this study, we introduced SaudiBERT, a new BERT-based language model pretrained from scratch exclusively on Saudi dialectal text. We evaluated the performance of SaudiBERT alongside other multidialectal Arabic language models such as MARBERT and CAMeLBERT-DA on 11 tasks divided into two groups: sentiment analysis and text classification. The target tasks include sarcasm detection, gender identification, event detection, sentiment analysis, and identification of Saudi dialect and its local variants. The results demonstrate the effectiveness of transformer-based models in handling tasks expressed in Saudi dialect, showing the effectiveness of using a domain-specific model like SaudiBERT which outperformed other multidialectal models. Additionally, we introduced two new Saudi dialectal corpora: STMC and SFC that are considered the largest corpora of their kind. The results achieved on the evaluation tasks will serve as a benchmark for future work. Additionally, more NLP tasks related to Saudi dialect should be explored, such as named entity recognition (NER), question answering, natural language inference, stance detection, fake news and spam detection. The corpora and the language model introduced in this paper will serve as valuable resources for future work in different domains and fields such as linguistics, artificial intelligence, language processing, and cultural studies.

\backmatter








\section*{Declarations}




\subsection{Conflict of interest}
The author declares no conflict of interest.

\subsection{Availability of data and materials}
SaudiBERT model is publicly available on \url{https://huggingface.co/faisalq/SaudiBERT}. STMC dataset is available on \url{ https://huggingface.co/faisalq/STMC}, and SFC-mini is available on \url{ https://huggingface.co/faisalq/SFC-mini}. 

\subsection{Code availability}
The code and the results are available on \url{https://github.com/FaisalQarah/SaudiBERT}.

\bibliography{sn-bibliography}

\end{document}